\documentclass{article} 
\usepackage{nips13submit_e,times}
\usepackage{url}
\usepackage[leqno, fleqn]{amsmath}
\usepackage{amssymb}
\usepackage{qtree}
\usepackage[numbers]{natbib}

\usepackage{gb4e}
\newcommand{\natneg}{\mathbin{^{\wedge}}}
\def\ii#1{\textit{#1}}

\title{Can recursive neural tensor networks\\learn logical reasoning?}

\author{
Samuel R. Bowman \\
NLP Group, Dept. of Linguistics\\
Stanford University\\
Stanford, CA 94305-2150 \\
\texttt{sbowman@stanford.edu}
}

\nipsfinalcopy 

\begin{document}
\maketitle

\begin{abstract}
Recursive neural network models and their accompanying vector representations for words have seen success in an array of increasingly semantically sophisticated tasks, but almost nothing is known about their ability to accurately capture the aspects of linguistic meaning that are necessary for interpretation or reasoning. To evaluate this, I train a recursive model on a new corpus of constructed examples of logical reasoning in short sentences, like the inference of \ii{some animal walks} from \ii{some dog walks} or \ii{some cat walks}, given that dogs and cats are animals. This model learns representations that generalize well to new types of reasoning pattern in all but a few cases, a result which is promising for the ability of learned representation models to capture logical reasoning.
\end{abstract}
\section{Introduction}

Deep learning methods in NLP which learn vector representations for words have seen successful uses in recent years on increasingly sophisticated tasks \cite{collobert2008unified, socher2011semi, socher2013acl1, chen2013learning}. Given the still modest performance of semantically rich NLP systems in many domains---question answering and machine translation, for instance---it is worth exploring the degree to which learned vectors can serve as general purpose semantic representations. 
Much of the work to date analyzing vector representations for words (see \cite{baroni2013frege}) has focused on lexical semantic behaviors---like the similarity between words like \ii{Paris} and \ii{France}. Good similarity functions are valuable for many NLP tasks, but there are real use cases for which it is necessary to know how words relate to one another or to some extrinsic representation, rather than just how similar they are to one another: deciding whether an inference holds for instance, such as that \ii{John went to Paris} means that \ii{John went to France} but not vice versa. This paper explores the ability of linguistic representations developed using supervised deep learning techniques to support interpretation and reasoning. 

Natural language inference (NLI), the ability to reason about the truth of a statement on the basis of some premise, is among the clearest examples of a task that requires comprehensive and accurate natural language understanding \cite{maccartney2009natural}. I borrow the structure of the task from \citet{maccartney2009natural}. In it, the model is presented with a pair of sentences, and made to label the logical relation between the sentences as equivalence, entailment, or any of five other classes, as here:

\begin{quote}
\ii{At least three dogs bark} \textbf{entails} that \ii{at least two dogs bark}.\\
\ii{All dogs bark} is \textbf{equivalent} to \ii{no dogs don't bark}.\\
\ii{Most dogs bark} is \textbf{incompatible} with \ii{no dogs bark}.
\end{quote}

The success of an inference system most crucially depends on the use of a representation of phrase and sentence meaning that supports the accurate computation of inferential consequences. The framework of monotonicity inference \cite{Hoeksema:1986, valencia91categorial} delimits a means of computing inferences in a broad set of cases, and I focus in this paper on reproducing this kind of inference in a learned model.

NLI involves many aspects of sentence understanding, and it is useful to be able to isolate inference from all the other ways that sentence understanding can go wrong: even if a model can learn to perform inference correctly, a failure of coreference resolution, word sense disambiguation, or parsing could supply it with inaccurate input and cause it to fail on an  evaluation. To sidestep this potential problem, I propose a simplified variant of the task below that uses hand-constructed, unambiguous inference problems, and construct a dataset of these problems.

In the following sections, I present the task of strict unambiguous NLI and present a recursive neural tensor network model configured for the task but closely modeled after those used in practice by \citet{socher2013acl1}. I then describe a dataset on which to evaluate the model, and lay out several experiments to test its ability to perform inference and to generalize to patterns of inference different from those seen at training time. I present results that reveal that the model is able to learn to reason using familiar patterns of inference with perfect accuracy and that it can generalize successfully to unseen patterns in many cases. I conclude with a discussion of the implications of these largely positive results, a brief error analysis, and proposals for future work.

\section{Reasoning with monotonicity}

Consider the statement \ii{all dogs bark}. From this, one can infer quite a number of other things. One can replace the first argument of \ii{all} (the first of the two predicates following it, here \ii{dogs}) with any more specific category that contains only dogs and get a valid inference: \ii{all puppies bark}; \ii{all collies bark.} But one can't replace it with anything more general: it doesn't give one reason to say that \ii{all animals bark}. 
The second argument of \ii{all} works the opposite way---one can make it less specific, but not more---one can say \ii{all dogs make sounds}, but one can't say that \ii{all dogs bark at cars}. These inferences depend at least partially on the choice of the quantifier \ii{all}. If we replace it with \ii{most}, the first argument doesn't work the same way: \ii{most dogs bark} doesn't entail that \ii{most animals bark} nor does it entail that \ii{most schnauzers bark}.

Monotonicity is the formalization of observations of this kind. The quantifier \ii{all} is downward monotone in its first argument because it permits the substitution of more specific terms, and upward monotone in its second argument because it permits the substitution of more general terms.
Formally, monotonicity is a property of certain positions in semantic structures that determines what kinds of substitution can be made in that position while preserving truth.
Most quantifiers allow monotonicity reasoning in some direction for both arguments. \ii{No}, for example is downward in both arguments, \ii{some} and \ii{two} are upward in both, and \ii{most} is upward in its second argument, but (unusually) does not license any kind of monotonicity inference on its first.

Monotonicity is a central insight from work on natural logic \cite{valencia91categorial, maccartney2009natural}, a theoretical framework for natural language inference that uses natural language strings as the logical symbols, rather than relying on conversion to and from first order logic or a similar system. The experiments here are not a direct implementation of natural logic, or even of monotonicity, but I rely on these theories to define what types of inference I include in the data, and in turn, what types of inference I expect the model to learn. 

I limit myself in this study to examples of reasoning involving quantifiers like \ii{some} and \ii{all}. They are of interest here for the plethora of monotonicity reasoning patterns that they license, but quantifiers as lexical items are of particular interest in vector space models for other reasons as well.
It is easy to imagine how a vector space might encode entities as vectors, with similar entities clustering together and certain directions in the space representing certain relationships between entities, and this interpretation of the dimensions of the space representing properties of entities can be expanded only somewhat awkwardly to common nouns and single-argument verbs---\citet{mikolov2013distributed} shows exactly this sort of behavior in VSMs learned for language modeling. This kind of similarity behavior is much less useful for quantifiers which define abstract relations between sets of entities: it might be somehow informative to encode lexical items such that \ii{two} and \ii{three} are more similar to each other than either is to \ii{no}, but this is not especially helpful for any task that involves interpreting or reasoning with these words.

\subsection{The task: natural language inference}

Natural language inference provides what I claim to be the simplest way to evaluate the ability of learned representation models to capture specific semantic behaviors. In the standard formulation of the task (and the one used in the RTE datasets \cite{dagan2006pascal}), the goal is to determine whether a reasonable human would infer a hypothesis from a premise.
MacCartney formalizes a method of inferring entailment relations, and moves past two way entailment/non-entailment classification, proposing the set $\mathfrak{B}$ of seven labels meant to describe all of the possible non-trivial relations that might hold between pairs of statements, shown in Table \ref{b-table}. 

\begin{table}
\begin{center}
\begin{tabular}{|c|c|c|c|} \hline
name & symbol & set-theoretic definition & example \\ \hline \hline
entailment & $x \sqsubset y$ & $x \subset y$ & \ii{crow, bird}  \\ \hline
reverse entailment & $x \sqsupset y$ & $x \supset y$ & \ii{Asian, Thai}  \\ \hline
equivalence & $x \equiv y$ & $x = y$ & \ii{couch, sofa} \\ \hline
alteration & $x$ $|$ $y$ & $x \cap y = \emptyset \wedge x \cup y \neq \mathcal{D}$ & \ii{cat, dog} \\ \hline
negation & $x \natneg y$ & $x \cap y = \emptyset \wedge x \cup y = \mathcal{D}$ & \ii{able, unable} \\ \hline
cover & $x \smallsmile y$ & $x \cap y \neq \emptyset \wedge x \cup y = \mathcal{D}$ & \ii{animal, non-ape} \\ \hline
independence & $x$ \# $y$ & (else) & \ii{hungry, hippo}\\ \hline
\end{tabular}
\caption{The entailment relations in  $\mathfrak{B}$. $\mathcal{D}$ is the universe of possible objects of the same type as those being compared, and the relation \# applies whenever none of the other six do, including when there is insufficient knowledge to choose a label.}
\label{b-table}
\end{center}
\end{table}

Interest in NLI in the NLP community is ongoing, and a version of it has been included in the 2014 SemEval challenge specially targeted towards the evaluation of distributional models. Neither this dataset, nor any other existing RTE and NLI datasets are appropriate for the task at hand however. I intend for the present experiment to evaluate the ability of a class of models to learn certain types of inference behavior, and need a dataset that precisely tests these phenomena. With existing datasets that use unrestricted natural language, there is the risk that a model could successfully capture monotonicity inference, but fail to accurately label test data due to problems with, for example, lexical ambiguity, syntactic ambiguity, coreference resolution, or pragmatic enrichment.

In order to minimize this possibility, I define the task of strict unambiguous NLI (SU-NLI). In this task, only entailments that are licensed by a strictly literal interpretation of the provided sentences are considered valid, and several constraints are applied to the language to minimize ambiguity:
\begin{itemize}
\ex A small, unambiguous vocabulary is used.
\ex All strings are given an explicit unlabeled tree structure parse.
\ex Statements involving the hard-to-formalize generic senses of nouns---i.e. \ii{dogs bark} as opposed to the non-generic \ii{all dogs bark}---are excluded.
\ex The sentences do not contain context dependent elements. This includes any reference to time or any tense morphology, and all pronouns.
\ex The morphology is dramatically simplified: the copula is not used (\ii{some puppy is French} is simplified to \ii{some puppy French}, to make it more directly comparable to sentences like \ii{some puppy bark}), and agreement marking (\ii{they walk} vs. \ii{she walks}) is omitted.
\end{itemize}

The key to success at this task is to learn a set of representations and functions that can mimic the logical structure underlying the data. There is limited precedent that deterministic logical behavior can be learned in supervised deep learning models: \citet{socher2012semantic} show in an aside that a Boolean logic with negation and conjunction can be learned in a minimal recursive neural network model with one-dimensional (scalar) representations for words. Modeling the logical behavior that underlies linguistic reasoning, though, is a substantially more difficult challenge, even in modular hand-built models.

The natural logic engine at the core of MacCartney's \cite{maccartney2009natural} NLI system requires a complex set of linguistic knowledge, much of which takes the form of what he calls projectivity signatures. These signatures are tables showing the entailment relation that must hold between two strings that differ in a given way (such as the substitution of the argument of some quantifier), and are explicitly provided to the model
for dozens of different cases of insertion, deletion and substitution of different types of lexical item. For example in judging the inference \ii{no animals bark $|$ some dogs bark} it would first used stored knowledge to compute the relations introduced by each of the two differences between the sentences. Here, the substitution of \ii{no} for \ii{some}  yields $\natneg$ and the substitution of \ii{dogs} for \ii{animals} yields $\sqsupset$. It would then use an additional store of knowledge about relations to resolve the resulting series of relations into one ($|$) that expresses the relation between the two sentences being compared:
\begin{quote}

1. \ii{no animals bark $\natneg$ \textbf{some} animals bark}\\
2. \ii{some animals bark $\sqsupset$ some \textbf{dogs} bark}\\
3. \ii{no animals bark $[\natneg\bowtie\thinspace\sqsupset\thinspace = |]$ some dogs bark}

\end{quote}

This study is the first that I am aware of to attempt to build an inference engine based on learned vector representations, but two recent projects have attempted to introduce vector representations into inference systems in other ways: 
\citet{baroni2012entailment} have achieved limited success in building a classifier to judge entailments between one- and two-word phrases (including some with quantifiers), though their vector representations were crucially based on distributional statistics and were not  learned for the task.
In another line of research, \citet{garrette2013formal} propose a way to improve standard discrete NLI with vector representations. They propose a deterministic inference engine (similar to MacCartney's) which is augmented by a probabilistic component that evaluates individual lexical substitution steps in the derivation using vector representations, though again these representations are not learned, and no evaluations of this system have been published to date.
\label{sec2}
\section{Methods}

The model is centered on a recursively applied composition function, following \citet{socher2011dynamic}, which is meant to mimic the recursive construction of meanings in formal models of semantics. In this scheme, pairs of words are merged into phrase representations by a function that maps from representations of length $2N$ to representations of length $N$. These phrase representations are then further merged with other words and phrases until the entire phrase or sentence being evaluated is represented in a single vector. This vector is then used as the input to a classifier and used in a supervised learning task.

\begin{figure}
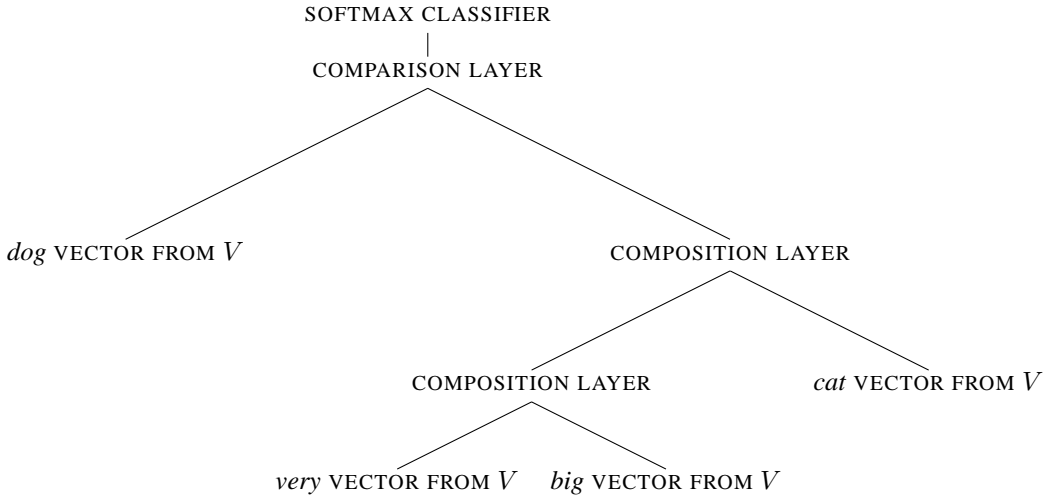

\begin{center}
\Tree [.{\sc softmax classifier}  [.{\sc comparison layer} [.{\sc \ii{dog} vector from $V$} ] [.{\sc composition layer} [.{\sc composition layer} [.{\sc \ii{very} vector from $V$} ] [.{\sc \ii{big} vector from $V$} ] ] [.{\sc \ii{cat} vector from $V$} ] ] ] ]
\end{center}
\caption{The model structure used to compare \ii{dog} and \ii{(very big) cat}. \label{sample-figure}} 
\end{figure}

Borrowing a model directly from the existing literature for this task is impossible since none has been proposed to detect asymmetric relations between phrases (though it may be possible to slightly adapt the paraphrase model of \citet{socher2011dynamic} to the task in future work). Instead, I build a combination model, depicted in Figure \ref{sample-figure}. The two phrases being compared are built up separately on each side of the tree using the same composition function until they have each been reduced to single vectors. Then, the two phrase vectors are fed into a separate comparison layer that is meant to generate a feature vector capturing the relation between the two phrases. The output of this layer is then given to a softmax classifier, which in turn produces a hypothesized distribution over the seven relations.

For a composition layer, I use the RNTN layer function proposed in \citet{chen2013learning} (itself adapted from \citet{socher2013acl1}) and shown below. The standard $\tanh$ sigmoid nonlinearity is applied to the output, following Socher et al.
\begin{equation} 
y_i = f_{a}(\vec{x}^{(l)T} \mathbf{A^{[i]}} \vec{x}^{(r)} + \vec{B_{i,:}} [\vec{x}^{(l)}; \vec{x}^{(r)}] + c_i)
\end{equation}
The comparison layer uses the same type of function with different parameters and a different nonlinearity function wrapped around it:
\begin{equation}
y_i = f_{b}(\vec{x}^{(l)T} \mathbf{K^{[i]}} \vec{x}^{(r)} + \vec{L_{i,:}} [\vec{x}^{(l)}; \vec{x}^{(r)}] + m_i)
\end{equation}
Rather than use a sigmoid here, I found a substantial improvement in performance by using a rectified linear function for $f_{b}$. In particular, I use the leaky rectified linear function \cite{maasrectifier}: $f_{b}(\vec{x})=\max(\vec{x}, 0) + 0.01\min(\vec{x}, 0)$,  applied elementwise. 

To run the model forward and label a pair of phrases, the structure of the lower layers of the network is assembled so as to mirror the tree structures provided for each phrase. The word vectors are then looked up from the vocabulary matrix $V$ (one of the model parameters), and the composition and comparison functions are used to pass information up the tree and into the classifier at the top. The model is trained using backpropagation through structure  \cite{goller1996learning}, wherein the negative log of the probability assigned to the correct label is taken as a cost function, and the gradient of each parameter with respect to that cost function is computed at each node, with information passing down the tree and into both the function parameters and the vocabulary matrix. Gradients for different instances of the composition function or different instances of a word in the same tree are summed to produce at most a single gradient update per parameter.

\textit{Optimization:} I train the model with stochastic gradient descent (SGD), with gradients pooled from randomly chosen minibatches of 32 training examples, and learning rates computed using AdaGrad \cite{duchi2011adaptive} from a starting rate of 0.2. L-BFGS \citep{nocedal1980updating} was tried unsuccessfully as an alternative to SGD. An L2 regularization term is added to the gradients from each batch, with a $\lambda$ coefficient of 0.0002. L1 regularization was tested but showed no improvement, though it may be fruitful in future work to explore other ways to encourage the learning of sparse solutions in the hope that they might be able to better support discrete, deterministic behavior of the sort studied here. The parameters and word vectors are initialized randomly using a uniform distribution over $[-0.1, 0.1]$---I made some attempts to initialize the word vectors using pretraining on a corpus of relations between individual vocabulary items (e.g. \ii{dog $\sqsubset$ animal}), but this offered no measurable improvement.

\textit{Model size:} The dimensionalities of the model were tuned extensively. The performance of the model is approximately optimal with the dimensionality of the word vectors set to 16, and the dimensionality of the feature vector produced by the comparison layer set to 45.

I also experimented with increasing the size of the model in two other ways: I ran all of the experiments below with an additional one or two standard neural network layers positioned between the comparison layer and the softmax classifier at the top of the network. This did not yield a consistent measurable improvement. I additionally reran the experiments with a variation of the network that uses different composition functions to compose different types of phrase, following \cite{socher2013acl}. This variation yielded no  improvement in performance over the standard RNTN model described here. More detail is provided in Appendix A.

\ii{Source code and data are available at} \url{http://goo.gl/PSyF5u}.

\section{Data}

The dataset is built on a small purpose-built vocabulary, which is intended to be large and diverse enough to permit a wide variety of experiments on different semantic phenomena beyond those shown here, but which is still small enough to allow the relations between the lexical items themselves to be exhaustively manually labeled when needed. The vocabulary contains 41 predicates (nouns, adjectives, and intransitive verbs; see Appendix B for more on the design of the wordlist), six quantifiers, the logical conjunctions \emph{and} and \emph{or}, and the unary operator \emph{not}.

The data take the form of 12k pairs of short sentences, each annotated with a relation label, which I divide into about 200 smaller datasets. These datasets contain tightly constrained variants of a specific valid reasoning pattern, each differing in at most one lexical item, and all sharing the same relation label. The remainder of this section describes the composition of these 200 datasets. Examples from four of the datasets are provided in Table \ref{examplesofdata}.

\begin{table}\small\centering
\begin{tabular}{|ll|l|}
\hline
\textbf{1. (some $x$) mobile $\sqsupset$ (some $y$) mobile [s.t. $x\sqsupset y$]}&~&\textbf{3. (all $x$) French $\sqsubset$ (some $x$) European}		
\\\hline
(some dog) mobile	 $\sqsupset$  (some puppy) mobile	&&
(all puppy) French $\sqsubset$ (some puppy) European
\\
(some animal) mobile $\sqsupset$ (some cat) mobile&&
(all cat) French $\sqsubset$ (some cat) European 		\\
	
(some Asian) mobile $\sqsupset$ (some Thai) mobile&&(all hippo) French $\sqsubset$ (some hippo) European \\ ...&&...\\\hline
\textbf{2. (all $x$) bark $\sqsubset$ (some $x$) bark} &&\textbf{4. (all $x$) bark $\equiv$ (no $x$) (not bark)}\\\hline
(all puppy) bark $\sqsubset$ (some puppy) bark	 	&&(all puppy) bark $\equiv$ (no puppy) (not bark)\\ 
(all cat) bark $\sqsubset$ (some cat) bark	&&(all cat) bark $\equiv$ (no cat) (not bark)\\ 
 (all hippo) bark $\sqsubset$ (some hippo) bark&&(all hippo) bark $\equiv$ (no hippo) (not bark)\\
 ...&&...\\\hline
\end{tabular}
\caption{Sample datasets from four different classes.\label{examplesofdata}}
\end{table}

\noindent\textit{Basic monotonicity:} This set of datasets contains reasoning patterns like those seen in example 1  in Table \ref{examplesofdata}, where one of the predicates on one side entails the predicate in the corresponding position on the other side. In some of the datasets (as in 1), this alternation is in the first argument position, in some the second argument position, and in some both. For the alternating first argument subclasses, I have every lexical entailment pair in the data in the first argument position, the terms \ii{bark}, \ii{mobile}, and \ii{European} in the second argument position, and every quantifier. 
For alternating second argument datasets, I have all predicates in the first argument position, and the pairs \ii{bark--animate}, \ii{French--European}, and \ii{Parisian--French} in the second argument position. The datasets in which there is an alternation in both positions fall into two categories. In some, the entailment relations between the arguments work in the same direction (\ii{some dog French} $\sqsubset$ \ii{some animal European}), in others, they work in opposite directions (\ii{some dog European} \# \ii{some animal French}).

\noindent\textit{Quantifier substitution:} These datasets, exemplified in 2 above, show the interactions between each possible pair of different quantifiers, with the same predicates on both sides. Datasets exist with \ii{bark}, \ii{mobile}, and \ii{European} in the second argument position, and within each dataset every possible predicate is shown in first argument position.

\noindent\textit{Monotonicity with quantifier substitution:} These datasets show the more complex monotonicity interactions that emerge when reasoning between phrases with differing arguments and differing quantifiers, as in  3. 
Exhaustively creating datasets of this kind involves too large an amount of data to readily verify manually, so I sampled  from the extensive set of possible combinations of the six quantifiers and the fixed argument fillers used in the monotonicity datasets. Each possible relation but `$\equiv$' (which does not apply sentences like these unless the predicates on both sides are identical) is expressed, as is every possible pair of quantifiers.

\noindent\textit{Negation:}
To show the interaction of negation and monotonicity, I included variants of many of the datasets described above in which one of the four argument positions is systematically negated, as demonstrated in example 4. 

\section{Experiments and results}

In the simplest experimental setting, which I label \textsc{all-split}, I randomly sample 85\% the data---making sure to sample 85\% of each of the individual datasets---train the model on that portion, and evaluate on the remaining 15\% of the data. This setting is meant to test whether the model is able to correctly generalize the individual reasoning patterns represented by each of the datasets. 

Performance on this setting is perfect: the model quickly converges to 100\% accuracy on the test data, showing that it is capable of accurately learning to capture all of the reasoning patterns in the data. The remaining experimental settings serve to determine whether what is learned captures the underlying logical structure of the data in a way that allows it to accurately label unseen kinds of reasoning pattern. In each of them, I choose one of three arbitrarily chosen target datasets, all involving quantifier substitution, to test on. I then then hold out that dataset and---depending on the experimental setting---other similar datasets from the training data in an attempt to discover just how different a test example can be from anything seen in training and still be classified accurately. Table \ref{patterntable} shows what information is included in the training data for each of the four settings for one of the three target datasets. 

\begin{table}\small\centering
\begin{tabular}{|l|cccc|}\hline
Dataset & \textsc{all-split} & \textsc{set-out} & \textsc{subcl.-out} & \textsc{pair-out} \\\hline
1. (some $x$) bark $\sqsupset$ (some $y$) bark [$y$ $\sqsubset$ $x$] &\checkmark&\checkmark&\checkmark&\checkmark\\
2. (all $x$) bark $\equiv$ (no $x$) (not bark)&\checkmark&\checkmark&\checkmark&\checkmark\\
3. (most $x$) bark $\#$ (no $y$) bark [$y$ $\sqsubset$ $x$]&\checkmark&\checkmark&\checkmark&\\
4. (no $x$) bark $\sqsupset$ (most (not $x$)) bark&\checkmark&\checkmark&\checkmark&\\
5. (most $x$) European $|$ (no $x$) European &\checkmark&\checkmark&&\\
6. (most $x$) bark $|$ (no $x$) bark &\checkmark&&&\\\hline
\end{tabular}
\caption{Examples of which datasets are included in the training data in each experimental setting for the target evaluation dataset \ii{(most $x$) bark $|$ (no $x$) bark}.  A `\checkmark' indicates that 85\% of the data from that dataset is seen in training, and a blank indicates that none of the examples from the dataset are seen in training. \label{patterntable}}
\end{table}

\begin{table}\small\centering
\begin{tabular}{|l|l|lll|}\hline
Target dataset & Data evaluated   & \textsc{set-out} & \textsc{subcl.-out} & \textsc{pair-out} \\\hline
\ii{(most $x$) bark $|$ (no $x$)} bark&target dataset only		&100\%&		100\%&	93.6\%\\
& all held out datasets                                			 		&(100\%) & 	36.8\%& 	78.8\%\\
& all test data                                 				 		&99.8\% & 	95.9\%&	93.8\%\\\hline
\ii{(two $x$) bark \# (all $x$) bark}&target dataset only  			&0\%&		100\%&		94.7\%\\
& all held out datasets                                			 		&(0\%) & 		100\%& 	62.7\%\\
& all test data                                  						&97.5\% & 	99.3\%&	93.0\%\\\hline
 \ii{(some $x$) bark $\natneg$ (no $x$) bark}&target dataset only &0\%&			0\%&		0\%\\
& all held out datasets                                 					&(0\%) & 		0\%& 	25.2\%\\
& all test data                                   						&97.7\% & 	94.0\%&	85.5\%\\\hline
\end{tabular}
\caption{Test set accuracy.\label{resultstable}}
\end{table}

\textsc{set-out:} In these experiments, I hold out the target dataset, training on none of it and testing on all of it, and additionally split each remaining dataset as in \textsc{all-split}. This setting is meant to test whether the model can generalize a broader reasoning pattern from one dataset to another---the model will still have seen other similar quantifier substitution datasets that differ from the target dataset only in which filler word is placed in the second argument position, as in row 5 in Table \ref{patterntable} above.

\textsc{subclass-out:} In these experiments, I hold out the target dataset as well as all of datasets representing the same reasoning pattern as the target dataset, and split each remaining dataset as in \textsc{all-split}. For the target dataset \ii{(most $x$) bark $|$ (no $x$) bark} in this setting, all of the datasets of the general form \ii{(most $x$) $y$ $|$ (no $x$) $y$} would be held out.

\textsc{pair-out:} In these experiments, I hold out all of the datasets that demonstrate the interaction between a particular pair of quantifiers. For the target dataset \ii{(most $x$) bark $|$ (no $x$) bark} in this setting, all examples with \ii{most} on one side and \ii{no} on the other, in either order, with or without negation or differing predicates, are held out from training. The remaining datasets are split as above.

These last three experiments have multiple sources of test data: some from the 15\% test portions of the datasets that were split, some from the target datasets, and some from any other similar datasets that were held out. In  Table \ref{resultstable}, I report results for each experiment on the entire test dataset, on the single target dataset, and on all of the held out datasets (i.e., the test data excluding the 15\% portions from the training datasets) in aggregate. This third figure is identical to the second for the \textsc{set-out} experiments, since the only held out dataset is the target dataset.

Performance was perfect for at least some held out datasets in both the \textsc{set-out} and \textsc{subclass-out} settings, and near perfect for some in \textsc{pair-out}. Performance for one of the three target datasets---\ii{(some $x$) bark $\natneg$ (no $x$) bark}---was consistently poor across training settings. This poor performance is consistent across random initializations, and should be a target of further investigation.

\section{Discussion} 

The model learns to generalize well to novel data in most but not all of the training configurations. This inconsistent performance suggests that there is room to improve the optimization techniques used on the model, but the fact that it is able to perform well in these settings even some of the time suggests that the structure of the model is basically capable of learning meaning representations that support inference.

The results in the \textsc{all-split} condition show two important behaviors. The model is able to learn to identify the difference between two unseen sentences and consistently return the label that corresponds to that difference. In addition, the model can learn lexical relations from the training data, such as \ii{dog $\sqsubset$ animal} from \ii{(no dog) bark $\sqsupset$ (no animal) bark}, and it can then use these learned lexical relations to compute the relation for a sentence pair like \ii{(some dog) bark $\sqsubset$ (all animal) bark}. The results from the other three experimental settings show that the model is able to learn general purpose representations for quantifiers which, at least in many cases, allow it to perform inference when a crucial difference between two sentences---the  substitution of one specific quantifier for another---has not been seen. These results serve to confirm that the representations learned are capturing some of the underlying logic, rather than than just supporting the memorization of fixed reasoning patterns. 
 
MacCartney's natural logic provides a perspective with which to view these results. That logic is not directly implemented here, and it is not meant to describe the behavior of a learned model, but it can provide some insight about what types of inference an ideal model should be able to learn to perform.
For example, one can use that logic to infer the unseen relation \ii{(most x) y $|$ (no x) y} by way of seen examples like \ii{(most x) y $\sqsubset$  (some x) y} and \ii{(some x) y $\natneg$ (no x) y} using a similar method to the one outlined in at the end of Section \ref{sec2}. 
 The quantifier pair \ii{some--no} and the target examples involving that pair behave somewhat unusually in this logic, which could shed some light on the model's failure to handle them well. If no examples with \ii{some} and \ii{no} are included in training, the logic doesn't have the information that it needs to derive the strict negation relation \ii{(some x) y $\natneg$ (no x) y} from the remaining data in these experiments. It can only instead derive the strictly weaker (less informative) relation $|$, by way of pairs of seen examples like \ii{(some x) y $\sqsupset$ (most x) y} and \ii{(most x) y $|$ (no x) y}. The learned model does exactly this, guessing $|$ for $\natneg$ in the target dataset 100\% of the time in the \textsc{set-} and \textsc{subclass-out} settings and 84\% of the time in the \textsc{pair-out} setting.
It is possible, then, that the model is capable of consistently learning general purpose representations for quantifiers in the \textsc{pair-out} setting, but only if the held out pair of quantifiers is one whose relation can be inferred from the training data under something like MacCartney's logic.
 
These results leave open the question of how much information is minimally needed to learn general purpose representations for quantifiers in this setting. There are two lines of experimental work that could help to clarify this. Including longer sentences and constructions involving conjunctions (i.e. \ii{and, or}), transitive verbs (i.e. \ii{eats, kicks}) or other constructions in the training and test data could further test what kinds of behavior can be learned from a small training set, as could further experiments on this data involving even smaller training sets than those shown here or more elaborately structured configurations of train and test sets.
In a more formal direction, a thorough study of MacCartney's system of projectivity and its underlying logic (following \citet{Icard:2012}) could lead to a clearer theoretical bound on what can be expected from learned models. Such formal work could also help to establish what tasks can be accomplished using vector representation models, and what tasks---if any---require more powerful word and phrase representations like the higher order tensors studied in \citet{coecke2010mathematical} and \citet{grefenstette2013towards}.

\subsubsection*{Acknowledgments}

I owe thanks to Chris Manning and Chris Potts for their advice at every stage of this project, to Richard Socher and Jeffrey Pennington for extensive and helpful discussions, and to J.T. Chipman for help in designing a pilot experiment.

\bibliographystyle{unsrtnat}

\bibliography{MLSemantics} 

\begin{thebibliography}{22}
\providecommand{\natexlab}[1]{#1}
\providecommand{\url}[1]{\texttt{#1}}
\expandafter\ifx\csname urlstyle\endcsname\relax
  \providecommand{\doi}[1]{doi: #1}\else
  \providecommand{\doi}{doi: \begingroup \urlstyle{rm}\Url}\fi

\bibitem[Collobert and Weston(2008)]{collobert2008unified}
Ronan Collobert and Jason Weston.
\newblock A unified architecture for natural language processing: {D}eep neural
  networks with multitask learning.
\newblock In \emph{Proc. {ICML}}. ACM, 2008.

\bibitem[Socher et~al.(2011{\natexlab{a}})Socher, Pennington, Huang, Ng, and
  Manning]{socher2011semi}
Richard Socher, Jeffrey Pennington, Eric~H. Huang, Andrew~Y. Ng, and
  Christopher~D. Manning.
\newblock Semi-supervised recursive autoencoders for predicting sentiment
  distributions.
\newblock In \emph{Proc. {EMNLP}}. {ACL}, 2011{\natexlab{a}}.

\bibitem[Socher et~al.(2013{\natexlab{a}})Socher, Perelygin, Wu, Chuang,
  Manning, Ng, and Potts]{socher2013acl1}
Richard Socher, Alex Perelygin, Jean~Y. Wu, Jason Chuang, Christopher~D.
  Manning, Andrew~Y. Ng, and Christopher Potts.
\newblock Recursive deep models for semantic compositionality over a sentiment
  treebank.
\newblock In \emph{Proc. EMNLP}. EMNLP, 2013{\natexlab{a}}.

\bibitem[Chen et~al.(2013)Chen, Socher, Manning, and Ng]{chen2013learning}
Danqi Chen, Richard Socher, Christopher~D. Manning, and Andrew~Y. Ng.
\newblock Learning new facts from knowledge bases with neural tensor networks
  and semantic word vectors.
\newblock In \emph{Proc. {ICLR}}, 2013.

\bibitem[Baroni et~al.(2013)Baroni, Bernardi, and Zamparelli]{baroni2013frege}
Marco Baroni, Raffaella Bernardi, and Roberto Zamparelli.
\newblock Frege in space: A program for compositional distributional semantics.
\newblock \emph{Submitted, draft at http://clic. cimec. unitn. it/composes},
  2013.

\bibitem[Mac{C}artney(2009)]{maccartney2009natural}
Bill Mac{C}artney.
\newblock \emph{Natural language inference}.
\newblock PhD thesis, Stanford University, 2009.

\bibitem[Hoeksema(1986)]{Hoeksema:1986}
Jack Hoeksema.
\newblock Monotonicity phenomena in natural language.
\newblock \emph{Linguistic Analysis}, 16\penalty0 (1--2), 1986.

\bibitem[S{\'a}nchez-Valencia(1991)]{valencia91categorial}
V{\'\i}ctor S{\'a}nchez-Valencia.
\newblock Categorial grammar and natural reasoning.
\newblock ILTI Publication Series for Logic, Semantics, and Philosophy of
  Language LP-91-08, University of Amsterdam, 1991.

\bibitem[Mikolov et~al.(2013)Mikolov, Yih, and Zweig]{mikolov2013distributed}
Tomas Mikolov, Wen-tau Yih, and Geoffrey Zweig.
\newblock Linguistic regularities in continuous space word representations.
\newblock \emph{Proceedings of NAACL-HLT}, 2013.

\bibitem[Dagan et~al.(2006)Dagan, Glickman, and Magnini]{dagan2006pascal}
Ido Dagan, Oren Glickman, and Bernardo Magnini.
\newblock The {PASCAL} {R}ecognising {T}extual {E}ntailment {C}hallenge.
\newblock In \emph{Machine Learning Challenges. Evaluating Predictive
  Uncertainty, Visual Object Classification, and Recognising Tectual
  Entailment}. Springer, 2006.

\bibitem[Socher et~al.(2012)Socher, Huval, Manning, and Ng]{socher2012semantic}
Richard Socher, Brody Huval, Christopher~D. Manning, and Andrew~Y. Ng.
\newblock Semantic compositionality through recursive matrix-vector spaces.
\newblock In \emph{Proc. {EMNLP}}, 2012.

\bibitem[Baroni et~al.(2012)Baroni, Bernardi, Do, and
  Shan]{baroni2012entailment}
Marco Baroni, Raffaella Bernardi, Ngoc-Quynh Do, and Chung-chieh Shan.
\newblock Entailment above the word level in distributional semantics.
\newblock In \emph{Proc. {EACL}}. {ACL}, 2012.

\bibitem[Garrette et~al.(2013)Garrette, Erk, and Mooney]{garrette2013formal}
Dan Garrette, Katrin Erk, and Raymond Mooney.
\newblock A formal approach to linking logical form and vector-space lexical
  semantics.
\newblock \emph{Computing Meaning}, 4, 2013.

\bibitem[Socher et~al.(2011{\natexlab{b}})Socher, Huang, Pennington, Ng, and
  Manning]{socher2011dynamic}
Richard Socher, Eric~H. Huang, Jeffrey Pennington, Andrew~Y. Ng, and
  Christopher~D. Manning.
\newblock Dynamic pooling and unfolding recursive autoencoders for paraphrase
  detection.
\newblock \emph{NIPS}, 24, 2011{\natexlab{b}}.

\bibitem[Maas et~al.(2013)Maas, Hannun, and Ng]{maasrectifier}
Andrew~L. Maas, Awni~Y. Hannun, and Andrew~Y. Ng.
\newblock Rectifier nonlinearities improve neural network acoustic models.
\newblock In \emph{Proc. ICML 30}, 2013.

\bibitem[Goller and Kuchler(1996)]{goller1996learning}
Christoph Goller and Andreas Kuchler.
\newblock Learning task-dependent distributed representations by
  backpropagation through structure.
\newblock In \emph{Proc. {IEEE} International Conference on Neural Networks},
  volume~1. IEEE, 1996.

\bibitem[Duchi et~al.(2011)Duchi, Hazan, and Singer]{duchi2011adaptive}
John Duchi, Elad Hazan, and Yoram Singer.
\newblock Adaptive subgradient methods for online learning and stochastic
  optimization.
\newblock \emph{The Journal of Machine Learning Research}, 2011.

\bibitem[Nocedal(1980)]{nocedal1980updating}
Jorge Nocedal.
\newblock Updating quasi-{N}ewton matrices with limited storage.
\newblock \emph{Mathematics of computation}, 35\penalty0 (151), 1980.

\bibitem[Socher et~al.(2013{\natexlab{b}})Socher, Bauer, Manning, and
  Ng]{socher2013acl}
Richard Socher, John Bauer, Christopher~D. Manning, and Andrew~Y. Ng.
\newblock Parsing with compositional vector grammars.
\newblock In \emph{Proc. ACL}, 2013{\natexlab{b}}.

\bibitem[Icard(2012)]{Icard:2012}
Thomas~F. Icard.
\newblock Inclusion and exclusion in natural language.
\newblock \emph{Studia Logica}, 100\penalty0 (4), 2012.

\bibitem[Coecke et~al.(2010)Coecke, Sadrzadeh, and
  Clark]{coecke2010mathematical}
Bob Coecke, Mehrnoosh Sadrzadeh, and Stephen Clark.
\newblock Mathematical foundations for a compositional distributional model of
  meaning.
\newblock \emph{arXiv:1003.4394}, 2010.

\bibitem[Grefenstette(2013)]{grefenstette2013towards}
Edward Grefenstette.
\newblock Towards a formal distributional semantics: Simulating logical calculi
  with tensors.
\newblock \emph{arXiv:1304.5823}, 2013.

\end{thebibliography}

\section*{Appendix A: The syntactically untied model}

I additionally ran some the experiments described above using a potentially more powerful variant of the model, following \citet{socher2013acl}, with three separately-parametrized composition functions instead of a single universal function.

All instances of the composition layer use the same basic RNTN layer structure, including the sigmoid nonlinearity, but the parameters are chosen from one of three sets as follows:

\begin{itemize}
\item \textit{Negation composition parameters:} Used in composing \ii{not} with any predicate, as in \ii{not $+$ European} or \ii{not $+$ dog}.
\item \textit{Quantifier first argument parameters:} Used in composing a quantifier with its first argument (a single word or a two-word phrase with \ii{not}), as in \ii{all} $+$ \ii{dog} or \ii{some} $+$ \ii{(not European)}.

\item \textit{Quantifier second argument parameters:} Used in composing a quantifier phrase (a quantifier with its first argument) with the quantifier's second argument (a single word or a two-word phrase with \ii{not}), as in \ii{all dog} $+$ \ii{bark} or \ii{most sofa} $+$ \ii{not indestructible}.

\end{itemize}

\section*{Appendix B: Wordlist}

\begin{tabular}{p{0.3\linewidth} p{0.3\linewidth} p{0.3\linewidth}}
hippo&hungry&non-ape\\
ape&animal&dog\\
cat&unable&able\\
Thai&Asian&bird\\
crow&sofa&couch\\
snail&vertebrate&invertebrate\\
bark (\ii{i.e. barks})&puppy&feline\\
mobile&immobile&dead\\
live&European&French\\
Parisian&human&animate\\
inanimate&seat&bench\\
woody&plant&tree\\
oak&bush&destructible\\a
indestructible&mammal&\\
\end{tabular}

This wordlist is part of a larger ongoing experiment, and is designed such that each of the seven of the relation types holds between some pair of words (for example, \ii{animal $\smallsmile$ non-ape}). Most of these relations are not included in the data used for this experiment. Since I do not train the model on a list of pairs of single words, lexical knowledge only comes in indirectly through monotonicity reasoning examples.
For example, if the model sees
\ii{(some dog) bark $\sqsubset$ (some animal) bark}, it could infer from that that \ii{dog $\sqsubset$ animal}. 

I avoid specifying most of the possible relations between words by any means to avoid giving the model so much background knowledge that it can evaluate whether a sentence it is trying to reason about is true. Though the task is defined such that the ground truth of the sentences being compared is irrelevant, mixing sentences that are known to be true with those that are not known to be true may provide unnecessary added difficulties in learning. For example, I do not train the model on \ii{all dog bark $\sqsubset$ all hungry bark} (thus teaching the model \ii{dog $\sqsubset$ hungry}), since this would make it potentially problematic to use later train the model on examples like \ii{some dog hungry $\sqsubset$ some animal hungry} where the left side is known to be true a priori.

\end{document}